\documentclass[10pt]{article}

\usepackage{tabularx}
\usepackage{multirow}
\usepackage[ruled]{algorithm2e}
\usepackage{amsmath}
\usepackage[pdftex]{xcolor,colortbl}
\usepackage{svg}
\usepackage[margin=10pt,font=small,labelfont=bf,labelsep=endash,textfont=it]{caption}
\usepackage{subcaption}
\usepackage{bm}
\usepackage{float}
\usepackage{amssymb}

\DeclareMathOperator*{\argmax}{arg\,max}

\begin{document}

\title{Classifications based on response times for detecting early-stage Alzheimer's disease}
\author{Alain Petrowski \\ RS2M, Telecom-SudParis, Institut Mines-Télécom\\ 9 rue Charles Fourier, 91011 Evry Cedex, France\\ Alain.Petrowski@telecom-sudparis.eu}
\date{}
\maketitle

\begin{abstract}
{\bf Introduction} --
This paper mainly describes a way to detect with high accuracy patients with early-stage Alzheimer's disease (ES-AD) versus healthy control (HC) subjects,  from datasets built with handwriting and drawing task records.

\medskip
{\bf Method} --
The proposed approach uses subject's response times. An optimal subset of tasks is first selected with a ``Support Vector Machine'' (SVM) associated with a grid search.
Mixtures of Gaussian distributions defined in the space of task durations are then used to reproduce and explain the results of the SVM. Finally, a surprisingly simple and efficient  ad hoc classification algorithm is deduced from the Gaussian mixtures.

\medskip
{\bf Results} --
The solution presented in this paper makes two or even four times fewer errors than the best results of the state of the art concerning the classification HC/ES-AD from the provided handwriting and drawing task records.

\medskip
{\bf Discussion} --
The best SVM learning model reaches a high accuracy for this classification but its learning capacity is too large to ensure a low overfitting risk regarding the small size of the dataset. The proposed ad hoc classification algorithm only requires to optimize three real-parameters. It should therefore benefit from a good generalization ability.

\end{abstract}




\section{Introduction}

Alzheimer's disease is the most common cause of dementia. The World Health Organization estimates the number of cases to be between 45 and 52 million worldwide in 2030 \cite{who2017}.  That is a major public health problem insofar as the number of patients is inexorably increasing, while the care required by each of them is heavy in the final stage of severe dementia. Alzheimer's disease is not yet well understood and the criteria for early diagnosis need to be refined.

\medskip

Authors have pointed out that the deterioration of motor skills, e.g. when walking or writing, often precedes the cognitive symptoms of Alzhei\-mer's disease by several years \cite{Buchman2011, Mielke2012}.
Writing involves fine motor skills that can be analyzed from the changes over time of pressure, position, altitude and azimuth of a pen on an electronic tablet.
Most of papers related to online handwriting analysis use that approach \cite{Impedovo2019, Kahindo2019}.
The identification of the patient groups or the healthy control group is achieved either using statistical tests or automatic classification methods \cite{Kahindo2019, Kahindo2018, El-Yacoubi2019}.
However, the best results published with such solutions reach 74\% \cite{El-Yacoubi2019} or 77\% \cite{Kahindo2018} accuracy, which is not truly satisfactory.

Another point of view is addressed in this paper. It does not take into account the movements of a pen.
Only the durations of a predefined set of handwriting or drawing tasks are used to classify a subject as healthy or having early-stage Alzheimer's disease (ES-AD).
That may be relevant since it has long been known that reaction or response times increase in ES-AD patients.
But this criterion is not specific to AD: other brain diseases and even the aging of healthy subjects also lead to increased response times \cite{Gordon1990}.

The machine learning approach described in this paper overcomes this difficulty by selecting an optimal set of tasks for which the response times allow a classifier to best recognize ES-AD cases. Compared to other methods applied to the dataset used for this paper, this results in the highest success rate achieved so far.

\medskip

The paper is organized as follows. Section \ref{sec:dataset} describes the dataset.
It is similar in size to the datasets used in many other contributions: 141 samples.
Section \ref{sec:SVM} details an automatic classifier based on Support Vector Machines (SVMs). They are used to recognize subjects as belonging to a healthy control group (HC), a group of patients with ES-AD or a third group with Mild Cognitive Impairment (MCI).
The best performances of the classifications HC/ES-AD, HC/MCI and MCI/ES-AD were assessed experimentally by exploring the space of  the best features and SVM hyperparameters with a grid search.

The interpretation of the results presented in section \ref{sec:SVM} led to the proposal of a probabilistic model in section \ref{sec:gaussian}, able to explain the good results obtained with SVMs.
An efficient very simple deterministic ad hoc classification algorithm described in section \ref{sec:algo} is then deduced from the probabilistic model.
Experiments have confirmed the similar performances of the SVMs and the probabilistic model compared to the ad hoc algorithm.
Finally, comparisons are made in section \ref{sec:comparisons} with results achieved from previous works using the same dataset.
Section \ref{sec:conclusion} concludes the paper.


\section{The dataset \label{sec:dataset}}

In cooperation with the geriatric unit of the Parisian hospital Broca and Telecom SudParis, the ALWRITE study has resulted in several publications with different methods to detect early-stage Alzheimer's disease, especially \cite{El-Yacoubi2019, Kahindo2018}. As part of this study, three groups of volunteer participants carried out various tasks related to writing, speaking and walking. 
\begin{itemize}
\item a Healthy Control group (HC): 27 participants,
\item a ``Mild Cognitive Impairment'' group (MCI) divided into 3 sub-groups:
  \begin{itemize}
  \item executive MCI (E-MCI): 42 participants,
  \item amnestic MCI (A-MCI): 7 participants,
  \item multi-domain MCI (MD-MCI): 38 participants,
  \end{itemize}
\item a group of patients with early-stage Alzheimer's disease (ES-AD): 27 participants.
\end{itemize}
The groups were determined by preliminary diagnoses for each of the participants.
The datasets built for this paper use responses from the participants of all groups, except the ``amnestic MCI'' group because its size is too small to yield meaningful results.


\subsection{Acquisition protocol}

We only consider 7 writing tasks among those of the ALWRITE project, simply because they were the only ones that were made by most of the 141 participants.

\begin{figure}[htb]
\begin{center}
\includegraphics[scale=0.26]{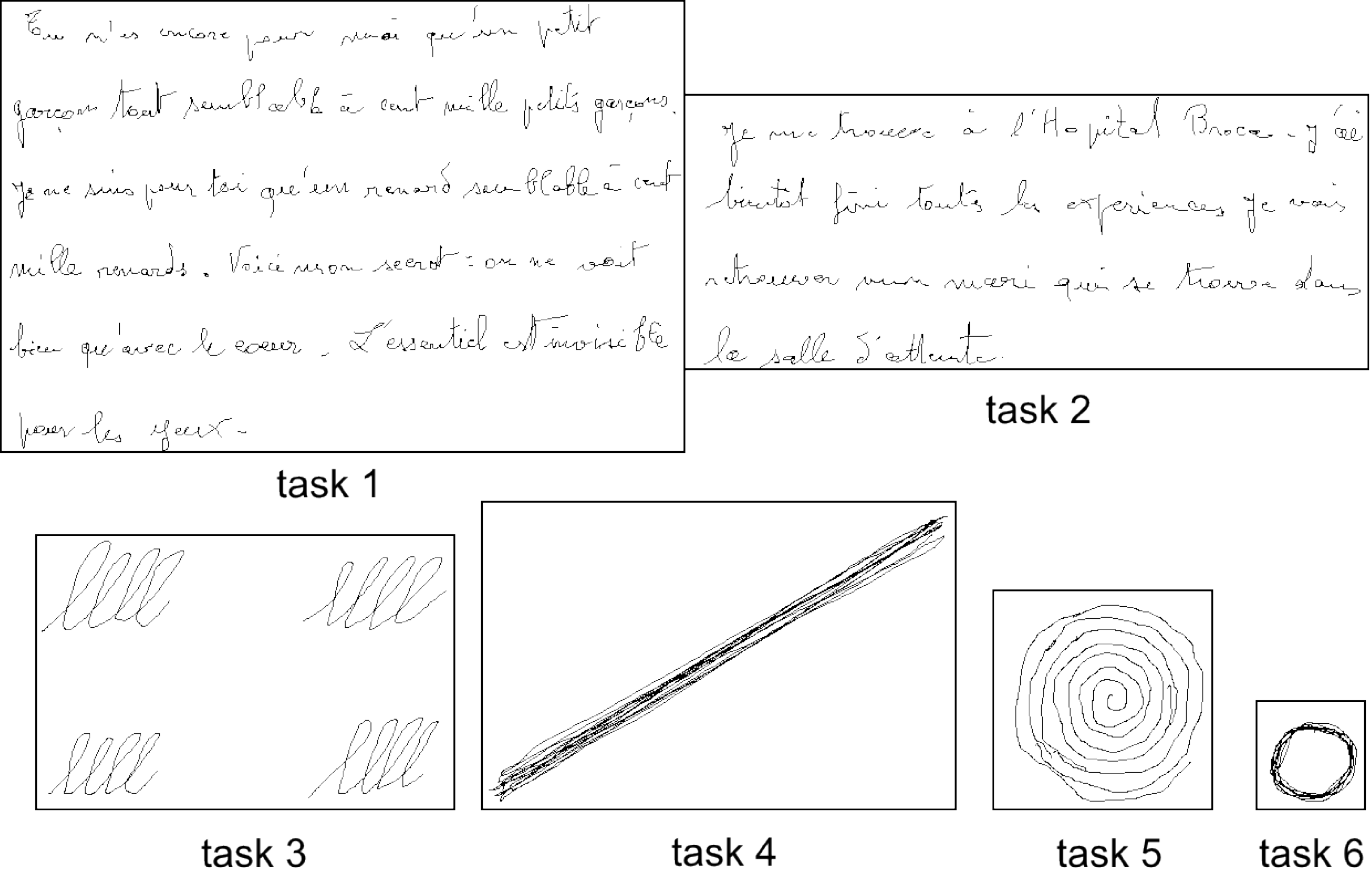}
\caption{An example of a response returned by one of the participants for the tasks of the acquisition protocol. }
\label{tasks}
\end{center}
\end{figure}

These tasks are listed below:
\begin{itemize}
\item task 1: copying a simple imposed text (see Fig. \ref{tasks});

\item task 2: writing a free text of four lines, preferably not memorized before;

\item task 3: writing four sets of four cursive $l$: $\ell\ell\ell\ell$ on given positions; the time during which the pen is on air is not significant;

\item task 4: Fitt's test: moving the pen back and forth between two given targets for 15 seconds;

\item task 5: drawing a spiral following a dotted line;

\item task 6: drawing circles on a given circle for 15 seconds;

\item task 7: static pen: holding the tip of the pen in a given position for 15 seconds.

\end{itemize}

However, the records for two participants in the ES-AD group are incomplete: tasks 4 to 7 are missing for one of them and task 5 is missing for the other one.
Likewise, a participant of the E-MCI group did not perform tasks 4 to 7.
This can reduce the number of participants depending on the sets of tasks considered in the experiments.

The participants used electronic pen tablets to record their responses under the supervision of an operator.
An example of such a response from a participant is given in Figure \ref{tasks}.
Those responses have been digitized as a table of numbers comprising 6 columns $(T, X, Y, P, Az, Al)$ and as many rows as there are measures.
Each row contains the instant $T$ in milliseconds of the measurement and at that time: the coordinates $X$ and $Y$ of the tip of the pen, its pressure $P$ on the tablet, its azimuth $Az$ and its altitude $Al$.

\subsection{Response time measurement\label{subsec:measurement}}

The {\em duration dataset} ${\bf S}$ is a set of samples ${s}_{k}, \forall k \in \{1, ..., |{\bf S}|\}$ each associated with a participant $k$.  $s_{k} = ({\bf f}_{k}, l_{k})$, where ${\bf f}_{k} = (f_{k1} ... f_{k7})$ is the {\it feature vector} of sample $k$ and $l_{k}$ is its label ``HC'', ``E-MCI'', ``A-MCI'', ``MD-MCI'' or ``ES-AD''.
The feature $f_{kj}$ is the duration of task $j \in \{1, ..., 7\}$ for sample $s_{k}$. 

It is quite natural to take into account the duration between the start and the end of a task to get response times from participants.
This is achieved by subtracting the time of the last non-zero pressure of the pen on the tablet from the time of the first non-zero pressure.
This choice is referred to as {\em measurement mode 1}. There are other possibilities to measure response times. Those listed below have been tried:

\begin{itemize}
\item mode 2: time during which the pen exerts a pressure on the tablet from the start to the end of a task: the length of time the pen is on air is not taken into account;

\item mode 3: the same as mode 1 except for task 3 (writing the loops $\ell$) for which mode 2 is used.

\item mode 4: like mode 3 except for tasks 4, 6 and 7 to be performed during an imposed time interval of 15 seconds. If those tasks last longer than 15 seconds, the response times are truncated to 15 seconds.

\item mode 5: this is the length of time the pen is on air from the start to the end of the tasks.

\end{itemize}


\section{Support Vector Machines for HC, MCI and ES-AD classifications \label{sec:SVM}}


\subsection{The classifier \label{sec:classifier}}

The results relative to this section have been obtained with ``Support Vector Machines'' (SVMs) \cite{Boser1992, James2013} to perform binary classifications between 
\begin{itemize}
\item HC and ES-AD groups,
\item HC and E-MCI groups,
\item HC and MD-MCI groups,
\item E-MCI and ES-AD groups,
\item MD-MCI and ES-AD groups.
\end{itemize}

These groups are of similar size, which gives hope for more significant classification results. SVMs are used because they are recognized as one of  the best binary classification methods. A machine learning model is often characterized by a set of hyperparameters whose values have to be determined before the training stage and remain constant during it.
Thus, the SVMs require to choose a regularization coefficient $c$ as well as a kernel function with its own hyperparameter value(s) \cite{James2013}.
The best choices depend on the datasets and there is no way to determine them a priori.
Radial basis function (RBF), linear and polynomial kernels \cite{James2013} were tested on the duration dataset. The RBF kernel gave the best performance and was chosen for this paper.
It is defined as:
$$
k({\bf x}, {\bf y}) = \exp(-\gamma ||{\bf x} - {\bf y}||^2)
$$
where ${\bf x}$ and ${\bf y}$ are feature vectors. $\gamma > 0$ is the only hyperparameter of the RBF kernel.

\;

The {\em grid search} is the simplest method, commonly used, to determine near optimal values of the hyperparameters.
It involves defining a small set of values for each hyperparameter.
Let $\bf C$ and $\bf \Gamma$ be sets of values for hyperparameters $c$ and $\gamma$ respectively.
The classifier is trained as many times as there are elements in ${\bf C} \times {\bf \Gamma}$.
The best hyperparameters in ${\bf C} \times {\bf \Gamma}$ are those for which the SVM minimizes, for example, the classification error rate on test sets.

\;

A model is also characterized by the subset of features selected from the dataset so that its performance is as high as possible.
Indeed, using all the features provided by the dataset as input to the classifier is not often satisfactory because that increases the risk of overfitting \cite{James2013, Goodfellow2016},  especially when the dataset size is small.
Instead, using feature subvectors can increase the classifier performance.

Let $p$ be the number of features of the dataset. A feature subvector can be defined with the help of an indicator vector ${\bf I}$ whose the $p$ components ${\rm I}_{j}$ belong to $\{0, 1\}$.
Then feature ${f}_{kj}, j \in \{1, ..., p\}$ is selected to be an input of the SVM if and only if ${\rm I}_{j} = 1$.
The grid search is then performed in set: ${\bf F} \times {\bf C} \times {\bf \Gamma}$, where ${\bf F}$ is the set of the $2^p - 1$ possible indicator vectors ${\bf I}$.

Finally, each feature is linearly scaled between $-1$ and $1$, as usual for SVMs, over all samples in the dataset so as not to degrade the performance of the classification \cite{Hsu2003}.


\subsection{Training, validation and test}

In this section, the focus is on the need to perform cross-validations and nested cross-validations when only a small dataset is available, to obtain the best possible machine learning model and to assess its generalization performance.

The model selection stage consists in finding the best values of the hyperparameters as well as selecting the best features to maximize the performance values on a validation set, e.g. with a grid search.
The validation set is built from samples which are not in the training set.
However, this method introduces an optimistic bias on the performance achieved on the validation set \cite{Hastie2001}.
The actual (unbiased) performance of the best model should then be assessed with a test set built from samples of the dataset that are neither in the training set nor in the validation set.
This model should avoid the overfitting risk by confirming the good performance of its predictions on the test set.
This risk is all the higher as the training set is small, all other things being equal.

Thus,  the dataset should be divided into 3 parts: a training set as large as possible to try to avoid overfitting,  a validation set  and a test set large enough to provide accurate estimates of performance. When a dataset is too small, such a method cannot be applied. Instead, we use {\em cross-validations} or else {\em nested cross-validations} when optimal hyperparameters and best selected features need to be determined \cite{Hastie2001}.


\subsubsection{Leave-One-Out cross-validation}

From a dataset $\bf S$ containing $n$ samples $s_{k}, \forall k \in \{1, ..., n\}$, the Leave-One-Out cross-validation (LOOCV) method builds $n$ training sets ${\bf S} \backslash \{s_{k}\} $ each of them containing $n - 1$ samples and $n$ test sets $\{s_k\}$ containing only one sample.
Its main advantage is that it reduces the size of the dataset by only one sample to build a training set.
In addition, the error evaluation $\cal E$ is a mean of the error ${\cal E}_{k}$ achieved during the $n$ trainings, thus improving the quality of the estimate of $\cal E$ compared to a simple hold-out validation on the same dataset.
At last, it is current to associate a cross-validation with feature selection and a tuning mechanism to optimize the hyperparameters, such a grid search, as shown in Algorithm \ref{algo:LOOCV}, in order to achieve the best performance from the machine learning algorithm.

\begin{algorithm}[h]
\DontPrintSemicolon 
\KwIn{${\bf S} = \{s_{1}, ..., s_{n}\}$: a dataset of $n$ samples $s_{i}$}
\KwOut{${\cal E}_{\rm best}$: minimal error,\;
\hspace*{1.5cm} ${\bf I}_{\rm best}$:  indicator vector of the best selected features,\;
\hspace*{1.5cm} $c_{\rm best}$, $\gamma_{\rm best}$: best hyperparameter values for the RBF SVM}
$\bf Variables:$\;
\hspace*{1.5cm} $c$, $\gamma$: hyperparameter values for the RBF SVM,\;
\hspace*{1.5cm} ${\bf I}$: indicator vector of selected features,\;
\hspace*{1.5cm} ${\bf F}, {\bf C}, {\bf \Gamma}$ are the sets of values of ${\bf I}$, $c$ and $\gamma$ respectively,
\;
${\cal E}_{\rm best} \gets 1$\;
\For(\textcolor{blue}{\tcp*[h]{grid search loop}}) {{\bf all} $({\bf I}, c, \gamma) \in {\bf F} \times {\bf C} \times {\bf \Gamma}$} {  
${\bf S'} \gets {\bf featureSelection}({\bf S}, {\bf I})$\;
\For(\textcolor{blue}{\tcp*[h]{cross-validation loop}}){$k \in \{1, ..., n\}$} {
  ${\cal M} \gets {\bf train}({\bf S'} \backslash \{s'_{k}\}, c, \gamma)$ {\textcolor{blue}{\tcp*[h]{${\cal M}$: model built by ${\bf train()}$}}} \;
  ${\cal E}_{k} \gets 
 \left\{
\begin{array}{cl}
0  & {\bf if }~{\bf predict}(s'_{k}, {\cal M}) = {\bf label}(s'_{k}) \\
1  &  {\rm otherwise}
\end{array}
\right.$ \;
 }
${\cal E} \gets  1 / n{{\sum_{k=1}^n {\cal E}_{k}}}$\;

\If {${\cal E} < {\cal E}_{\rm best}$} {
${\cal E}_{\rm best} \gets {\cal E}$\;
${\bf I}_{\rm best}, c_{\rm best}, \gamma_{\rm best} \gets {\bf I}, c, \gamma$\;
}
}
\Return{${\cal E}_{\rm best}$, ${\bf I}_{\rm best}, c_{\rm best}, \gamma_{\rm best}$ }\;

\caption{ Leave-One-Out cross-validation associated with a grid search and a feature selection}
\label{algo:LOOCV}
\end{algorithm}


\subsubsection{Nested cross-validation}

However, the unbiased performance assessment needs to use a test set different from the training and validation sets when hyperparameter tuning and feature selection are used.
Solutions have been proposed to meet this requirement, such that the ``nested cross-validation'' (NCV) algorithm and its variants \cite{Stone1974}.

For the present paper, a version of NCV derived from LOOCV is used (Algorithm \ref{algo:LOONCV}).

\begin{algorithm} 
\DontPrintSemicolon 
\KwIn{${\bf S} = \{s_{1}, ..., s_{n}\}$: a dataset}
\KwOut{$a$: model assessment (estimated generalization accuracy),\;
$\bf Variables:$\;
\hspace*{1.5cm} ${\bf I}_{\rm best}$: indicator vector of the best selected features,\;
\hspace*{1.5cm} $c_{\rm best}$, $\gamma_{\rm best}$: best hyperparameter values for the RBF SVM,}
\hspace*{1.5cm} $c$, $\gamma$: hyperparameter values for the RBF SVM,\;
\hspace*{1.5cm} ${\bf I}$: indicator vector of selected features,\;
\hspace*{1.5cm} ${\cal M}$: SVM model,\;
\hspace*{1.5cm} ${\cal E}$: error count\;
\;
\For(\textcolor{blue}{\tcp*[h]{NCV outer loop}}){$i \in \{1, ..., n\}$} {
${\bf S}_{i} \gets {\bf S}\backslash \{s_{i}\}$\;
${\cal E}_{\rm best} \gets n$\;
\For(\textcolor{blue}{\tcp*[h]{grid search loop}}) {{\bf all} $({\bf I}, c, \gamma) \in {\bf F} \times {\bf C} \times {\bf \Gamma}$} {  
${\bf S'}_{i} \gets {\bf featureSelection}({\bf S}_{i}, {\bf I})$\;
${\cal E} \gets 0$\;
\For(\textcolor{blue}{\tcp*[h]{NCV inner loop}}){$k \in \{1, ..., n-1\}$} {
  ${\bf L} \gets {\bf S'}_{i} \backslash \{s'_{k}\}$\;
 
  ${\cal M} \gets {\bf train}({\bf L}, c, \gamma)$ \;
  ${\cal E} \gets {\cal E} +  \left\{
\begin{array}{cl}
0  & {\rm if}~{\bf predict}(s'_{k}, {\cal M}) = {\bf label}(s'_{k}) \\
1  &  {\rm otherwise}
\end{array}
\right.$ \; 
}

\If {${\cal E} < {\cal E}_{\rm best}$} {
${\cal E}_{\rm best} \gets {\cal E}$\;
${\bf I}_{\rm best}, c_{\rm best}, \gamma_{\rm best} \gets {\bf I}, c, \gamma$\;
}
}
  ${\bf S'}_{i} \gets {\bf featureSelection}({\bf S}_{i}, {\bf I}_{\rm best})$\;
  $s'_{i} \gets {\bf featureSelection}(s_{i}, {\bf I}_{\rm best})$\;
  ${\cal M} \gets {\bf train}({\bf S'}_{i}, c_{\rm best}, \gamma_{\rm best})$ \;
  $a \gets a +  \left\{
\begin{array}{cl}
1/n  & {\rm if }~{\bf predict}(s'_{i}, {\cal M}) = {\bf label}(s'_{i}) \\
0  &  {\rm otherwise}
\end{array}
\right.$ \; 
}
\Return{$a$}\;

\caption{The Nested Cross-Validation (NCV) algorithm used for this paper}
\label{algo:LOONCV}
\end{algorithm}

The nature of performance $a $ depends on the problem to be solved. Here, $a$ is the accuracy to maximize:
$$
a = {\sum_{i=1}^n{\delta(l_{i}, p_{i})} \over n}
$$
where $\delta(x, y) = 1$ if $x = y$, otherwise  $\delta(x, y) = 0$.  $l_{i}$ is the desired label for sample $i$ while $p_{i}$ is the label predicted by the model.


\subsection{Experiments and results\label{sec:experiments}}

The experiments aim to select and assess the best learning models for the HC/ES-AD, HC/E-MCI, HC/MD-MCI, E-MCI/ES-AD, MD-MCI/ES-AD classifications, according to the  duration measurement modes 1 to 5, the optimal hyperparameters and selected tasks.
The grid search is performed in the  hyperparameter space with:
\[
\begin{array}{c}
c \in {\bf C} = \{0.1, 0.2, 0.5, 1, 2, 5, 10, 20, 50, 100\}   \\
\gamma \in {\bf \Gamma} = \{0.1, 0.2, 0.5, 1, 2, 5, 10, 20, 50, 100\}
\end{array}
\]
and in the set of $2 ^7 - 1$ indicator vectors  ${\bf I}$ of the selected tasks: ${\rm I}_{k} = 1$ if the duration of task $k$ is selected to be a feature, otherwise: ${\rm I}_{k} = 0$.
The grid search must therefore explore $(2 ^7 - 1) \times |{\bf C}| \times |{\bf \Gamma}| = 12700$ possible configurations to find the best learning model for each classification.

For each configuration, the accuracies of learning models are assessed  by ``Leave One Out'' cross-validations (LOOCV) and nested cross-validations (NCV). LOOCV is used to find the best hyperparameters and feature selection, while NCV give estimates of accuracies without the optimistic bias of LOOCV.


\subsubsection{HC/ES-AD classifications \label{HC/ES-AD}}

\begin{table}[b] 
\caption{Classification HC/ES-AD: accuracies observed with Leave-One-Out cross-validations (CV) and nested cross-validations (NCV), according to the measurement modes of durations, the optimal hyperparameters $c_{\rm best}$ and $\gamma_{\rm best}$ as well as the best selected tasks.}
\begin{center}
\begin{tabular}{|c||c|c|c||c|c|c|}
\hline
 \multirow{2}{*}{mode}  &  best  & \multirow{2}{*}{$c_{\rm best}$}  & \multirow{2}{*}{$\gamma_{\rm best}$} & test with the  &        \multirow{2}{*}{CV} & \multirow{2}{*}{NCV} \\
            &   selected tasks  &         &                   & training set   &      ~       &  ~     \\
\hline
1 & 2, 4, 7         & 50  & 5   & 96\% & 91\%  & 79\% \\
2 & 1, 3, 4, 5, 7 & 50  & 2   & 98\% & 85\%  & 65\% \\
3 & 2, 4, 7         & 50  & 5   & 96\% & 91\%  & 79\% \\
{\bf 4} & {\bf 2, 4, 7}    & {\bf 0.5} & {\bf 10} & {\bf 94\%} & {\bf 94\%} & {\bf 94\%} \\
5 & 2, 3, 4, 6     & 20   & 10 & 90\% & 87\%  & 72\% \\
\hline
\end{tabular}
\end{center}
\label{table1}
\end{table}

The results are shown in Table \ref{table1}.
Mode 4 of the duration measurement gives by far the best performance for LOOCV: only 3 classification errors over 53 samples, corresponding to an accuracy of 94\% for selected tasks 2, 4, 7 and optimal hyperparameters $c = 0.5, \gamma = 10$.
For this configuration, the sensitivity is 96\% and the specificity is 93\%. The confusion matrix is given below with HC as the negative class and ES-AD as the positive class:
$$
\left(
\begin{array}{cc}
\text {true~positive } & \text {false~positive}   \\
\text {false~negative} & \text {true~negative}
\end{array}
\right) : 
\left(
\begin{array}{cc}
25  & 2  \\
1  & 25
\end{array}
\right)
$$

This optimal result in mode 4 is also observed with the nested cross-validation implemented according to Algorithm \ref{algo:LOONCV}.
For this mode, the optimal selected features and hyperparameters were chosen by all the folds of the NCV except the best value of $c$, which is found equal to 1 instead of 0.5 for only one fold over 53.
This stability is a sign of good quality learning, which also gives hope for a low overfitting.
For the other modes, the accuracies achieved with NCV are notably lower than those achieved by LOOCV.
The NCV shows instabilities of the learning models chosen by each fold: these models generalize poorly.

Column 5 of Table \ref{table1} also gives the accuracies when the 53 samples of the dataset are used as the training set and the test set.
For modes other than 4, the accuracies achieved with the training set as a test set are notably better than the accuracies with LOOCV: this is a clear sign of overfitting.
On the other hand, in mode 4, these accuracies are identical, equal to 94\%, which reinforces the hope for a low overfitting.
An accuracy of 94\% is a good result while the LOOCV and NCV cross-validations did not reveal any overfitting.
However, that best solution found by the SVM requires 43 support vectors out of 53 samples. When the number of support vectors is close to the size of the training set, this indicates that the risk of overfitting is a priori high.

\begin{figure} [h]
\begin{center}
\includegraphics[scale=0.3]{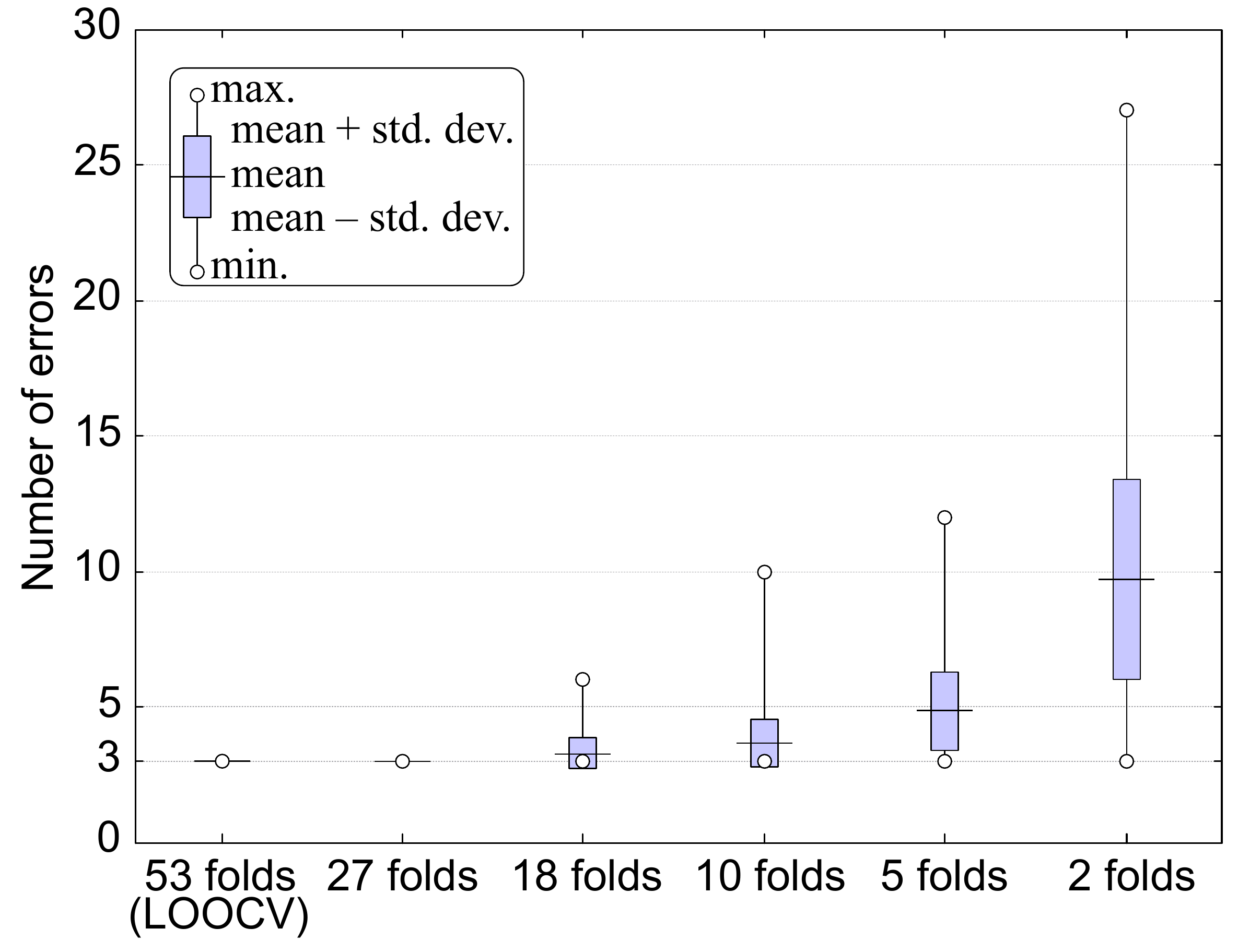}
\caption{HC/ES-AD classifications: distributions of the numbers of misclassification errors, according to the number of folds of the cross-validations, for selected tasks 2, 4 and 7 and with hyperparameters $c =  0.5$ and $\gamma = 10$.}
\label{fig:histoAccuracy}
\end{center}
\end{figure}

\paragraph{}
To further estimate the risk of overfitting, cross-validations have been implemented for 27, 18, 10, 5 and 2 folds.
Thus, several samples are removed from the training set according to the number of folds, which can degrade the performance of the machine learning models.
By randomly shuffling the samples from the dataset, many cross-validations have been performed for each fold number.
This made it possible to construct a histogram of accuracies for each number of folds.
The results are presented in Figure \ref{fig:histoAccuracy} for the optimal set of selected tasks 2, 4 and 7 and with the optimal SVM hyperparameters $c =  0.5$, $\gamma = 10$.

When 27 folds are used for 53 samples in the dataset, 2 samples are removed from the training sets for all but one of the folds.
Figure \ref{fig:histoAccuracy} shows that only 3 samples are misclassified for 100\% of the tests.
Thus, no degradation of performance occurs compared to LOOCV (see Table \ref{table1} and label ``53 folds'' in Figure \ref{fig:histoAccuracy}).
This good result is linked to the high accuracy achieved with NCV, because NCV also removes 2 samples from the dataset to make up the training set for each fold.

With 18-fold CVs, the size of the training sets is reduced by 3 samples.
A slight degradation of performance is observed since there are 3 to 6 misclassification errors with a mean of 3.3, instead of 3 errors with 27 folds.
This slight performance degradation when removing 3 samples from the training sets suggests that the risk of overfitting should be moderate when the number of samples in the training set is greater than 50.


\subsubsection{HC/{$x$}-MCI and {$x$}-MCI/ES-AD classifications}

The results of binary classifications between HC, ES-AD and MCI subgroups are given in Table \ref{table2}. 
The best nested-cross-validation accuracies of the HC/$x$-MCI classifications according to the measurement modes of durations are at least 19\% lower compared to the HC/ES-AD classification (94\%) in mode 4 (Table \ref{table1}).
\begin{table} [h!]
\caption{Classifications HC/{$x$}-MCI and {$x$}-MCI/ES-AD: accuracies observed with Leave-One-Out cross-validations (CV) and nested cross-validations (NCV), according to the measurement modes 1 to 5.}
\begin{center}
\begin{tabular}{|c||c|c||c|c||c|c||c|c|}
\hline
meas. & \multicolumn{2}{c||}{HC/E-MCI}  & \multicolumn{2}{c||}{HC/MD-MCI} & \multicolumn{2}{c||}{E-MCI/ES-AD} & \multicolumn{2}{c|}{MD-MCI/ES-AD} \\
mode &    CV     &  NCV &  CV            & NCV  &  CV                 & NCV  & CV                    & NCV \\
\hline
1 &    78\%  & 63\%       &  80\% & 60\%      &  ~86\%~ & 73\%          &  ~~83\%~~ & 63\% \\
2 &    79\%  & {\bf 75\%} &  74\%  & 48\%        &  88\% & 53\%          &  86\% & {\bf 78\%} \\
3 &    78\%  & 65\%         &  78\% & {\bf 69\%} &  86\% & 73\%          &  83\% & 60\% \\
4 &    75\%  & 43\%         &  78\% & 62\%        &  82\% & 61\%           &  86\% & 73\% \\
5 &    72\%  & 60\%         &  77\% & {\bf 69\%} &  85\% & {\bf 79\%}   &  75\% & 60\% \\
\hline
\end{tabular}
\end{center}
\label{table2}
\end{table}

The $x$-MCI/ES-AD classifications are better than HC/{$x$}-MCI ones but also clearly worse than HC/ES-AD one.
For all the best classifications with {$x$}-MCI according to the measurement modes of durations, the NCV performances are 4\% to 8\% lower than the CV performances, which indicates an overfitting. The confusion matrices for the best CV accuracies in Table \ref{table2} are given below with HC as negative class and ES-AD as positive class:
$$
\text{HC/E-MCI mode 2:}
\left(
\begin{array}{cc}
39  & 12  \\
2  & 15
\end{array}
\right)
~~~~~~~~~~\text{HC/MD-MCI mode 1:}
\left(
\begin{array}{cc}
35  & 10  \\
3  & 17
\end{array}
\right)
$$
$$
\text{E-MCI/ES-AD mode 2:}
\left(
\begin{array}{cc}
19  & 1  \\
7  & 40
\end{array}
\right)
~~~~~~\text{MD-MCI/ES-AD mode 2:}
\left(
\begin{array}{cc}
19  & 2  \\
7  & 36
\end{array}
\right)
$$


\subsection{Discussion\label{sec:discussion}}

A high accuracy of 94\% has been observed with a SVM classifier only for the HC/ES-AD classification when using the durations of task 2 (writing a free text), task 4 (Fitt's test) and task 7 (static pen) with measurement mode 4.

The errors made by HC/$x$-MCI and $x$-MCI/ES-AD classifications for the best result (from Table \ref{table2},  col. ``E-MCI/ES-AD NCV'', row ``mode 5'' : $100 - 79 = 21\%$) are at least 3.5 times more numerous than those of the HC/ES-AD classification (from Table \ref{table1}, col. ``NCV'', row ``mode 4'': $100 - 94 = 6\%$).

When tasks 2, 4 and 7 are selected to create feature vectors for the HC/ES-AD classification, with $c = 0.5$ and $\gamma = 10$, 43 support vectors for the SVM over 53 samples in the dataset suggests that the risk of overfitting is high a priori. But, the cross-validations show no evidence of overfitting with those parameters and the available dataset. A clear answer to the question of overfitting would require results with larger enough datasets. However, as additional data is not available, deepening this apparent contradiction requires more study than simple trainings of SVMs.

In the following sections, high-accuracy models with lower learning capacities than SVMs will be presented to show that the risk of overfitting is likely to be low.

\section{Gaussian mixtures for HC/ES-AD classification \label{sec:gaussian}}

In this section, the term ``dataset'' refers to the 53 HC and ES-AD samples, whose features vectors are the durations of tasks 4, 7 and 2 measured in mode 4.
From the results reported in the previous section, the aim of this section is to propose a simple, plausible and explainable probabilistic model for the HC/ES-AD classification. That classification is the only one considered because it gives the highest accuracy with SVMs.
The probabilistic model will be validated by the classification performance achieved with it.

\begin{figure}[h] 
\begin{center}
\includegraphics[scale=0.9]{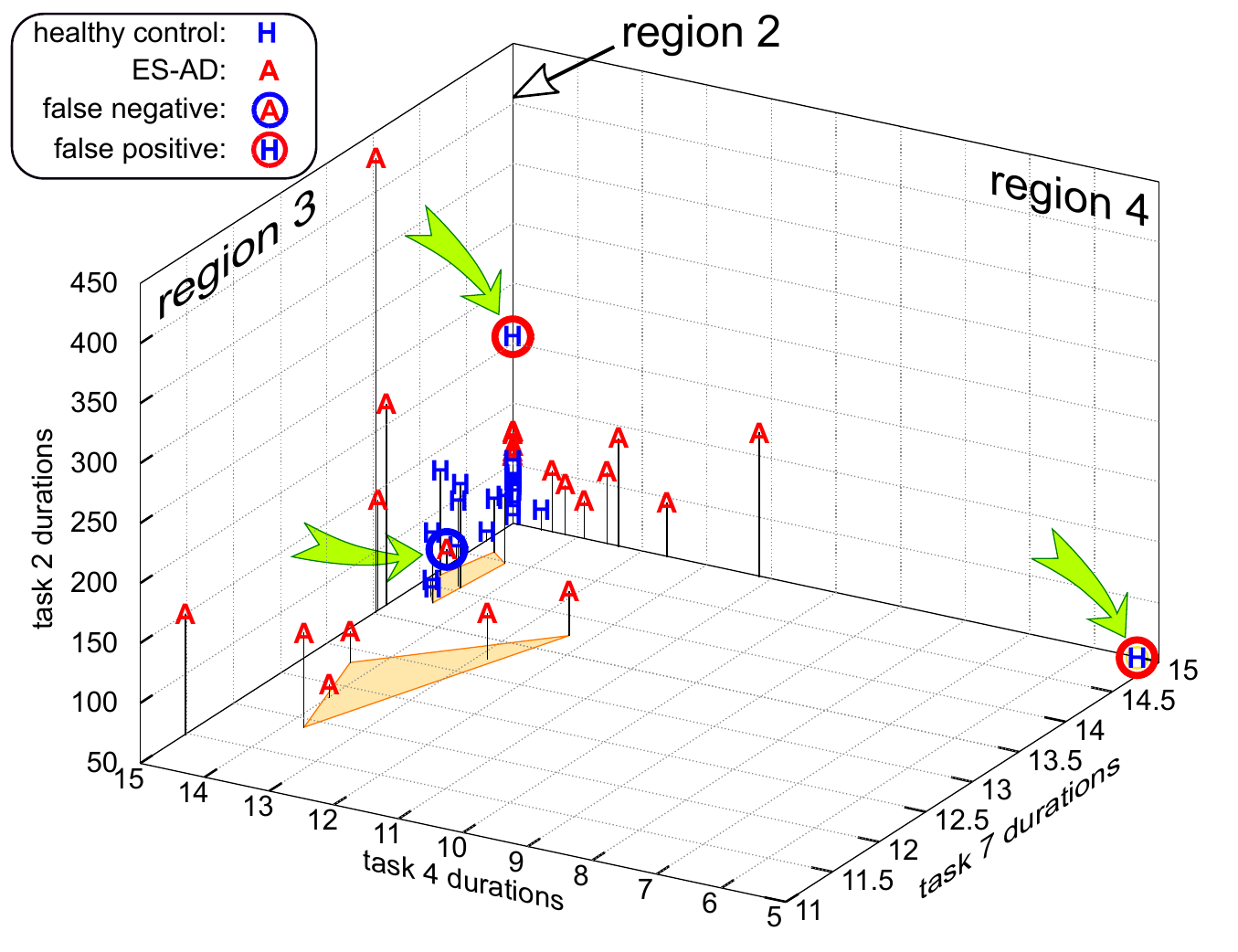}
\caption{Repartition of the HC and ES-AD samples in the space of durations for tasks 4, 7 and 2  in ``measurement mode 4'', to achieve the best accuracy of 94\% with the SVMs. Hyper-parameters: cost $c = 0.5$,  RBF kernel, $\gamma = 10$. The 3 misclassified samples are indicated by the green arrows.}
\label{ResponseTime247}
\end{center}
\end{figure}

Figure \ref{ResponseTime247} shows the distributions of the samples of the dataset in the 3D space generated by the axes $(t_{4}, t_{7}, t_{2})$ respectively associated with the durations of tasks 4, 7 and 2.
A sample is represented by the letter ``$\textcolor{blue}{\bf H}$'' if it is classified as ``Healthy Control'' or ``$\textcolor{red}{\bf A}$''  if it is classified as ``Early-Stage Azheimer's Disease'' with the SVMs.
A sample predicted as HC when it is labeled ES-AD in the dataset is represented by the letter ``$\textcolor{red}{\bf A}$'' surrounded by a blue circle, which means ``false negative''.
Similarly, a sample predicted as ES-AD when it is labeled HC in the dataset is represented by the letter ``$\textcolor{blue}{\bf H}$'' surrounded by a red circle, which means ``false positive''.

The feature space is divided into 4 regions to take into account the prescribed time of 15 seconds for tasks 4 and 7 in ``measurement mode 4'' :
\begin{itemize}

\item Region 1 contains the samples whose durations of tasks 4 and 7 do not exceed the time prescribed of 15 seconds.
It is a 3D box whose points belong to $[0, 15) \times [0, 15) \times [0, 450]$ in the coordinate system $(t_{4}, t_{7}, t_{2})$. 15 is the time prescribed in seconds for tasks 4 and 7, while 450 is an upper bound of observed durations for task 2 over all the samples in the dataset. The region contains 8 HC and 5 ES-AD samples of the dataset. The convex envelopes of the projections of those samples onto the $(t_{4}, t_{7})$  plane (orange areas) improve their visualization.

\item Region 2 is a straight line segment whose points are in $\{15\} \times \{15\} \times  [0, 450]$.  It is detailed in Figure \ref{fig:AxisAnalysis472}. Samples with durations of tasks 4 and 7 exceeding the time prescribed of 15 seconds are projected onto region 2. It contains the largest number of samples compared to the other regions: 15 HC and 9 ES-AD samples.

\item Region 3 is a rectangle whose points are in $\{15\} \times [0, 15) \times [0, 450]$. Samples for which only the durations of task 4 exceed the time prescribed of 15 seconds are projected onto region 3. This region contains 2 HC and 5 ES-AD samples.

\item Region 4 is a rectangle whose points are in $[0, 15) \times \{15\} \times [0, 450]$. Samples for which only the durations of task 7 exceed the time prescribed of 15 seconds are projected onto region 4. This region contains 2 HC and 7 ES-AD samples.

\end{itemize}


\subsection{The classifier}

Each region contains a set of samples for each of the two classes. For instance in Fig. \ref{ResponseTime247}, the two sets in region 1 for samples HC and ES-AD are highlighted by the convex envelopes (in orange) of their projections onto the plane $(t_{4}, t_{7})$.

From the above decomposition of the feature space into regions, the proposed probabilistic model is defined by 4 mixtures of 2 multivariate Gaussian distributions, each associated with a region $r$ and a label (or class) $c \in \{\text{HC}, \text{ES-AD}\}$.
The distributions are assumed to be Gaussian a priori because they are maximum entropy distributions in space $\mathbb{R}^n$ for a given vector of means and a given covariance matrix.
They are chosen uncorrelated because there is not enough data to assume otherwise, so the covariance matrices are diagonal, which simplifies matters.
Choosing such a simple model is an approximation which reduces the number of parameters to be optimized and therefore reduces the risk of overfitting.
If that approximation is not valid, it will result in a large bias error (underfitting) that will be evaluated by the experiments (in section \ref{sec:GM-Experiments}) \cite{Goodfellow2016}.
To respect the region boundaries, the Gaussian distributions are truncated between 0 and 15 for the $t_{4}$ and $t_{7}$ task durations.

\medskip

The mixture model gives the probability density $p_{r}({\bf x})$ to get an HC or ES-AD sample at any point $\bf x$ in region $r$ of the feature space:
\begin{equation} \label{eq:p(x)}
p_{r}({\bf x}) =  \sum_{c}\pi_{rc} {\cal N}({\bf x} | \bm{\mu}_{rc}, \bm{\sigma}_{rc}), \text{ ~~ } \forall r\,c, \pi_{rc} \geq 0 \text{ and }\sum_{c} \pi_{rc} = 1
\end{equation}
where $\pi_{rc}$ is the weight of the Gaussian distribution in region $r$ for label $c$;
 ${\cal N}({\bf x} | \bm{\mu}_{rc},  \bm{\sigma}_{rc})$ is the Gaussian density at $\bf x$;
 $\bm{\mu}_{rc}$ and $\bm{\sigma}_{rc}$ are respectively the vector of means and the vector of standard deviations deduced from the diagonal of the covariance matrix:

 \begin{equation} \label{eq:N(x)}
{\cal N}({\bf x} | \bm{\mu}_{rc}, \bm{\sigma}_{rc}) = \prod_{i=1}^{d_{r}}{1 \over \sqrt{2 \pi } \sigma_{rc,i}} \exp \left( -{1 \over 2} {\left(x_{i} - \mu_{rc,i} \over \sigma_{rc,i}\right)^2}\right)
\end{equation}
where $d_{r}$ is the number of non-zero standard deviations for region $r$: $d_{1} = 3$, $d_{2} = 1$, $d_{3} = d_{4} = 2$. $\sigma_{rc,i}$ is the $i$-th non-zero standard deviation for region $r$ and label $c$.

\medskip

The mixture weights $\pi_{rc}$ are estimated from the dataset: $\hat{\pi}_{rc} = n_{rc} / n_{r}$, where $n_{rc}$ is the number of samples in region $r$ for label $c$ and $n_{r} = \sum_{c} n_{rc}$ is the number of samples in the region. $\mu_{rc,i}$ and $\sigma_{rc,i}$ are also estimated from the dataset:

 \begin{equation} \label{eq:estim}
\hat{\mu}_{rc,i} = {1 \over n_{rc}}\sum_{j=1}^{n_{rc}}{\bf x}_{j,i}, ~~~~~~ \hat{\sigma}_{rc,i} = \sqrt{{1 \over n_{rc} - 1}\sum_{j=1}^{n_{rc}}({\bf x}_{j,i} - \hat{\mu}_{rc,i})^{2}}
\end{equation}
where ${\bf x}_{j}$ is a feature vector of the dataset in region $r$ for label $c$.

\medskip

These Gaussian mixtures (GMs) can be used to predict samples as HC or ES-AD in a more explainable way than the SVM learning models described in section \ref{sec:classifier}.
From this point of view, the training step consists in computing the parameters $\hat{\pi}_{rc}$, $\hat{\bm{\mu}}_{rc}$ and $\hat{\bm{\sigma}}_{rc}$ with eq. (\ref{eq:estim}) from a training set.

\medskip

Not all samples are used to compute the Gaussian parameters. Indeed, some samples are considered atypical because at least one of their task durations is farther than 4 standard deviations $\hat{\bm{\sigma}}_{rc}$ from its class mean $\hat{\bm{\mu}}_{rc}$. They are:
\begin{itemize}
\item in region 2: the misclassified HC sample for which the duration of task 2 is greater than 200 seconds (see Fig. \ref{ResponseTime247}),
\item in region 3: the well classified ES-AD sample for which the duration of task 2 is greater than 400 seconds,
\item in region 4: the misclassified HC sample for which the duration of task 4 is less than 6 seconds.
\end{itemize}
 Estimations of standard deviations and means of the Gaussian distributions are given in Table \ref{table:model1}.
Such a model needs 40 real-parameters to be defined: 8 weights $\pi_{rc}$, 8 vectors $\hat{\bm{\mu}}_{rc}$ and $\hat{\bm{\sigma}}_{rc}$ in $\mathbb{R}^{d_{r}}$.

\begin{table}[h]
\caption{Gaussian mixtures parameters to classify HC / ES-AD samples with the durations of tasks 4, 7 and 2.
The standard deviation for the HC samples cannot be estimated in region 4 because it contains only one of them.}
\begin{center}
\begin{tabular}{|c|c||c|c|c|c|}
\hline
region $r$ & label $c$    & $\hat{\pi}_{rc}$ & $(\hat{\mu}_{rc,4}~\hat{\mu}_{rc,7}~\hat{\mu}_{rc,2})^{\rm T}$ & $(\hat{\sigma}_{rc,4}~\hat{\sigma}_{rc,7}~\hat{\sigma}_{rc,2})^{\rm T}$\\
\hline
\hline
\multirow{2}{*}{1}   & HC            &  8/13  & $(14.6 ~ 14.1 ~ 103)^{\rm T}$     & $(0.16 ~ 0.27 ~~ 27)^{\rm T}$ \\
                              & ES-AD      &  5/13  & $(13.5  ~ 12.8 ~~~  88)^{\rm T}$       & $(0.8 ~~ 0.7 ~~~ 23)^{\rm T}$ \\
\hline
\multirow{2}{*}{2}    & HC$^{*}$  &  15/24  & $(15 ~~~~ 15 ~~~~ 82)^{\rm T}$ & $(0 ~~~~~ 0 ~~~~~ 12)^{\rm T}$ \\
                               & ES-AD      &  9/24    & $(15 ~~~~ 15 ~~~ 117)^{\rm T}$ & $(0 ~~~~~ 0 ~~~~\,~~ 9)^{\rm T}$ \\
\hline
\multirow{2}{*}{3}     & HC                &  2/7  & $(15 ~~~ 14.5 ~~~  59)^{\rm T}$      & $(0 ~~\,~ 0.16 ~~ 1.7)^{\rm T}$ \\
                                & ES-AD$^{*}$ &  5/7  & $(15 ~~~ 13.2  ~~ 144)^{\rm T}$    & $(0 ~~~\,~ 1.1 ~~~ 54)^{\rm T}$ \\
\hline
\multirow{2}{*}{4}     & HC$^{*}$ &  2/9       & $(14.5 ~~ 15 ~~~~ 67)^{\rm T}$         & $(- ~~~~ - ~~~~ -)^{\rm T}$ \\
                                & ES-AD     &  7/9       & $(13.3 ~~ 15 ~~~ 113)^{\rm T}$       & $(1.0 ~~~~ 0 ~~~~ 29)^{\rm T}$ \\
\hline
\multicolumn{5}{l}{$^{*}$ \footnotesize{including an atypical sample removed to calculate vectors $\hat{\bm{\mu}}_{rc}$ and $\hat{\bm{\sigma}}_{rc}$}\rule[-7pt]{0pt}{20pt}} 
\end{tabular}
\end{center}
\label{table:model1}
\end{table}%

For each feature vector $\bf x$ in a test set,  the testing step consists in comparing the densities $p_{rc}({\bf x})$ for the region $r$ that contains $\bf x$, with $c \in \{\text {HC}, \text{ES-AD}\}$, from eq. (\ref{eq:N(x)}) and (\ref{eq:predict}).
The predicted label $l({\bf x})$ is obtained  by choosing the one for which the density is maximum:

 \begin{equation} \label{eq:predict}
l({\bf x}) = \argmax_{c \in \{\text{HC}, \text{ES-AD}\}} p_{rc}({\bf x}) ~~~ \text{with} ~~~ p_{rc}({\bf x}) = \pi_{rc}\, {\cal N}({\bf x} | \bm{\mu}_{rc}, \bm{\sigma}_{rc})
\end{equation}


\subsection{Experiments and results \label{sec:GM-Experiments}}

The accuracies achieved with the GMs are given in Table \ref{tab:gaussian}. The ``SVMs: NCV'' column recalls the results given in Table \ref{table1} with nested cross-validations and durations measured in mode 4.
The 3 errors made with the SVMs are also made with the GMs.

\begin{figure} [h]
\begin{center}
\includegraphics[scale=0.7]{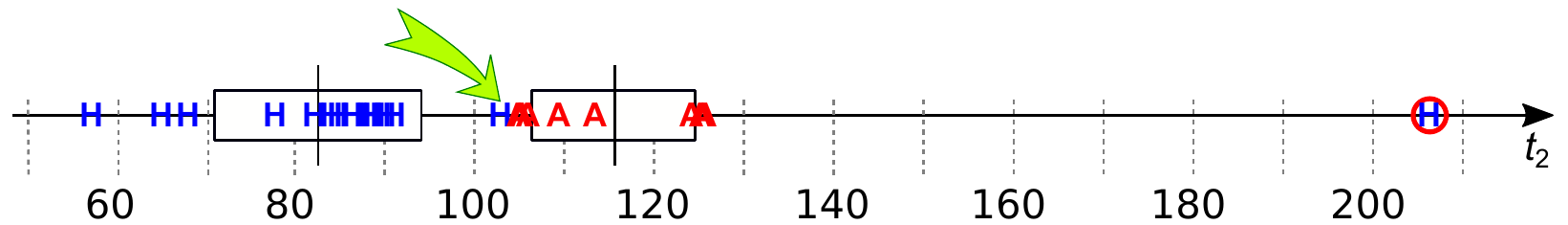}
\caption{Durations in seconds of task 2 for participants who exceeded the time prescribed of 15 seconds for tasks 4 and 7 (region 2).
The legend is the same as in Fig. \ref{ResponseTime247}.
The box-plots represent estimates of means and standard deviations.
The HC sample indicated by the green arrow is well classified by the SVMs but misclassified by the Gaussian mixtures.}
\label{fig:AxisAnalysis472}
\end{center}
\end{figure}

\begin{table}[htp]
\caption{Results of tests obtained with the Gaussian mixtures (GMs) on the HC/ES-AD dataset (53 samples) compared to those of nested cross-validations with Support Vector Machines (column  ``SVMs: NCV'').}
\begin{center}

\begin{tabular}{|c||c|c|c|}
\hline
~ &  SVMs: & GMs: training & GMs: \\
~ &  NCV    & set as the test set & LOOCV  \\
\hline
\hline
 error counts: &   3      &   4      &  7         \\
 \hline
 accuracies:   &  94\% & 92\%  &  87\%  \\
\hline
 specificities:   & 93\%  & 89\%   & 78\%    \\
\hline
 sensitivities:   & 96\%  & 96\%   & 96\%    \\
\hline
\end{tabular}
\end{center}
\label{tab:gaussian}
\end{table}%
The additional error observed with the GMs and the ``training set as the test set'' (column 3) is located in region 2 (indicated by the green arrow in Figure \ref{fig:AxisAnalysis472} ).
This is understandable since this HC sample is very close to the ES-AD cluster in region 2.

The Leave-One-Out Cross-Validation (LOOCV) for the GMs (column 4) makes three more errors. They only occur in regions 3 and 4 because they contain at most two HC samples.
Indeed, when an HC sample is removed to create a LOOCV fold, the single other HC sample in the region cannot give an estimate of a standard deviation.
These errors are actually caused by the too small dataset size.
However, the LOOCV accuracy shows no optimistic bias a priori. For this reason, it is more significant than the evaluation achieved with the ``training set as the test set''.


\subsection{Discussion \label{sec:GM-discussion}}

The hypothesis of a set of Gaussian mixtures (GMs) to explain the available dataset is confirmed by the achieved good results.
While being efficient, the GMs do not need hyperparameters such as $c$ and the kernel choice for the SVMs.
Table \ref{table:model1} shows that the behaviors of HC and ES-AD participants are clearly different.

\begin{figure} [h]
\begin{center}
\includegraphics[scale=0.75]{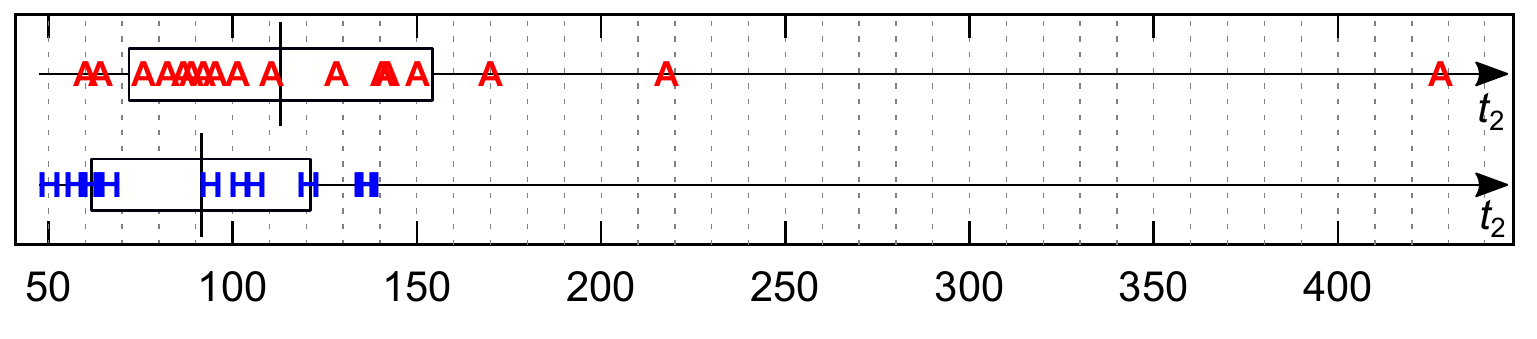}
\caption{Durations in seconds of task 2 for participants who responded at least once below 15 seconds for tasks 4 and 7  (union of regions 1, 3 and 4).
The box-plots represent estimates of means and standard deviations.
The legend is the same as in Fig. \ref{ResponseTime247}}.
\label{fig:AxisAnalysis134}
\end{center}
\end{figure}

\begin{itemize}
\item  For tasks 4 (Fitt's test) and 7 (static pen), the HC samples are close to the time prescribed of 15 seconds in regions 1, 3 and 4 (Fig. \ref{ResponseTime247}), except the atypical HC sample in region 4. Thus, most of HC samples in regions 1, 3 and 4 with durations $x_{4}$ and $x_{7}$ for tasks 4 and 7 respectively, are such that:
\begin{equation} \label{eq:p_4}
\exists p_{4} \in [0, 15), \exists p_{7} \in [0, 15) \mid x_{4} \geq p_{4} \text{ and } x_{7} \geq p_{7}
\end{equation}
where parameters $p_{4}$ and $p_{7}$ have to be determined to minimize the number of misclassified samples.

Furthermore, regions 3 and 4 contain only 2 HC samples, which is few compared to the other regions.
This would mean that most HC participants tend to have the same type of response time for tasks 4 and 7: either both durations are below the time prescribed of 15 seconds in region 1 or both durations are over beyond in region 2.

In contrast, the durations of tasks 4 and 7 for the ES-AD samples are more widely distributed than HC samples in regions 1, 3 and 4, as shown by the standard deviations $\hat{\sigma}_{rc,4}$ and $\hat{\sigma}_{rc,7}$ in Table \ref{table:model1}.
Thus, for the available dataset, it appears that ES-AD participants have less ability than HC participants to monitor their response time for simple tasks that do not require fine motor skills such as the ``Fitt's test'' and the ``static pen''.

\item For task 2 (writing a free text)  in region 2, the 99\% confidence intervals of the average response times of HC and ES-AD participants are respectively $[73, 92]$ and $[107, 127]$. They are calculated from Table \ref{table:model1}.
Based on these intervals, in average, ES-AD participants take between 16\% and 74\% longer than HC  participants to complete task 2.
It is remarkable that the sample standard deviations estimated for the two groups are small enough such that their response times are well separated  (Fig. \ref{fig:AxisAnalysis472}).
Thus, most of the HC samples in region 2 with duration $x_{2}$ for task 2 should be such as:
\begin{equation} \label{eq:p_2}
\exists p_{2} > 0 \mid x_{2} \leq p_{2}
\end{equation}
where $p_{2}$ has to be determined to minimize the number of misclassified samples.
The durations of task 2 (writing a free text) in region 2 are sufficient to clearly separate the HC and ES-AD samples.

In contrast, for task 2, by comparing between regions, Table \ref{table:model1} shows that the std. dev. in region 1 is more than twice the std. dev. in region 2: $\hat{\sigma}_{1,{\rm HC},2} > 2 \hat{\sigma}_{2,{\rm HC},2}$, $\hat{\sigma}_{1,{\rm ES-AD},2} > 2 \hat{\sigma}_{2,{\rm ES-AD},2}$. This is quite surprising because it would mean that the std. dev. of task 2 durations is higher when the durations of tasks 4 and 7 are below the prescribed time of 15 seconds. This observation can be generalized with the union of regions 1, 3, 4 vs. region 2, where $\hat{\sigma}_{134,{\rm HC},2} \approx 30$ and $\hat{\sigma}_{134,{\rm ES-AD},2} \approx 41$, as shown in Fig. \ref{fig:AxisAnalysis472} and \ref{fig:AxisAnalysis134}. If confirmed on other datasets, the reason might be neurological in nature.

Considering the union of regions 1, 3 and 4, Fig. \ref{fig:AxisAnalysis134} shows that the means of task 2 durations $\hat{\mu}_{134,{\rm HC},2} \approx 91 $ and $\hat{\mu}_{134,{\rm ES-AD},2} \approx 113$ are too close to each other compared to the high values of the std. dev. to separate classes HC and ES-AD with a high accuracy.
Thus, only durations of tasks 4 and 7 are available to separate HC and ES-AD participants in these regions.
In fact, tasks 4 and 7 are sufficient. That can be proven by removing the durations of task 2 from the dataset in regions 1, 3 and 4.
It is then observed after training that the accuracies given in Table \ref{tab:gaussian} are not altered.
\end{itemize}

\section{A simple prediction algorithm for HC/ES-AD classification \label{sec:algo}}


\subsection{The classifier}

From the above discussion (section \ref{sec:GM-discussion}) and results, as well as eq. (\ref{eq:p_4}) and (\ref{eq:p_2}), a very simple algorithm is proposed to predict the label associated with a feature vector (Algorithm \ref{algo:model}) whose the components are the durations of tasks 4, 7 and 2 measured in mode 4.
The duration threshold parameters $p_{4}, p_{7}, p_{2}$, associated with tasks 4, 7 and 2, have to be determined during training to minimize the number of errors, for example with a grid search.

\begin{algorithm} 
\DontPrintSemicolon 
\KwIn{${\bf x} = \{x_{4}, x_{7}, x_{2}\}$: durations of tasks 4, 7 and 2 (feature vector)\;
\hspace*{1.3cm}${\bf p} = \{p_{4}, p_{7}, p_{2}\}$: duration threshold parameters}
\KwOut{$l \in \{\text{HC},  \text{ES-AD}\}$: the predicted label or class for $\bf x$}\;

\uIf(\textcolor{blue}{~~~~~~~~\tcp*[h]{i.e.: if {\bf x} is in region 2}}){$x_{4} \geq 15$ {\bf and} $x_{7} \geq 15$} {
  $l \gets \text{HC {\bf if }}  x_{2} \leq p_{2} \text{ {\bf else} ES-AD}$\textcolor{blue}{~~~~\tcp*[h]{from eq. (\ref{eq:p_2})}}\;
}
\Else{ 
  $l \gets \text{HC {\bf if }}  x_{4} \geq p_{4} \text{ {\bf and} } x_{7} \geq  p_{7} \text{ {\bf else} ES-AD}$\textcolor{blue}{~~~~\tcp*[h]{from eq. (\ref{eq:p_4})}}\;
}
\Return{$l$}\;

\caption{A simple algorithm to predict the label HC or ES-AD from durations of tasks 4, 7 and 2.}
\label{algo:model}
\end{algorithm}


\subsection{Experiments and results}

The accuracies achieved with Algorithm \ref{algo:model} are given in Table \ref{tab:algo}. The ``SVMs: NCV'' column recalls the results given in Table \ref{table1} with the nested cross-validations and the durations measured in mode 4.

\begin{table}[H]
\caption{Results of tests obtained with Algorithm \ref{algo:model} on the HC/ES-AD dataset (53 samples) compared to those of nested cross-validations with Support Vector Machines (column “SVMs: NCV”)}
\begin{center}

\begin{tabular}{|c||c|c|c|}
\hline
~ &  SVMs: & Algo. \ref{algo:model}: training & Algo. \ref{algo:model}: \\
~ &  NCV    & set as the test set & LOOCV  \\
\hline
\hline
 error counts: &   3      &   4      &  6         \\
 \hline
 accuracies:   &  94\% & 92\%  &  89\%  \\
\hline
 specificities:   & 93\%  & 93\%   & 89\%    \\
\hline
 sensitivities:   & 96\%  & 92\%   & 88\%    \\
\hline
\end{tabular}
\end{center}
\label{tab:algo}
\end{table}

The 3 errors made with the SVMs are also made with Algorithm \ref{algo:model}.
When the test set is the training set (53 samples), the optimal parameter values are: $p_{4} \in [14.1, 14.3]$, $p_{7} = 13.7 \text{ and } p_{2} \in [103, 106]$ for a minimum number of errors equal to 4 and an accuracy of 92\%. It is equal to the performance obtained with the GMs (Table \ref{tab:gaussian}).

Two more errors are made by the Leave-One-Out Cross-Validation (LOOCV) of the algorithm for an accuracy of 89\%. This performance is lower but more significant than the test result achieved with the training set for the reasons already given in section \ref{sec:GM-Experiments}.


\subsection{Discussion}

First, this simple algorithm is more efficient than the GM model (section \ref{sec:gaussian}) from which it is derived, for the available dataset.

Also, from eq. (\ref{eq:p_4}), $x_{4} = p_{4}$ and $x_{7} = p_{7}$ are then the equations of two separator hyperplanes between classes HC and ES-AD samples in regions 1, 3 and 4. 
In a similar way, from eq. (\ref{eq:p_2}), $x_{2} = p_{2}$ is also the equation of a separator hyperplane between classes HC and ES-AD samples in region 2.

Thus, it only needs to adjust 3 real-parameters:  $p_{4}$, $p_{7}$ and $p_{2}$ during training, while being able to reach a good accuracy on a training set containing 53 times more numbers.
For this reason, overfitting is much less likely than with the SVMs (43 support vectors) or Gaussian models (40 real-parameters).


\section{Comparisons with previous works on the same dataset \label{sec:comparisons}}

The dataset described in section \ref{sec:dataset} was used in previous studies on the HC/ES-AD classifications \cite{El-Yacoubi2019, Kahindo2018, Kahindo2019}. The approaches chosen by the authors consist in particular in taking into account local velocities, accelerations and jerks of the tip of the pen.
The HC or ES-AD label is inferred from a Bayesian classifier. 
The best results were achieved with the local velocities for two tasks: writing an imposed text (task 1) p. 167 of \cite{Kahindo2019} and four sets of ``$\ell\ell\ell\ell$'' (task 3) \cite{Kahindo2018}. The authors compared them to experimental results with methods described in the literature. They have thus shown that their solutions outperform the state of the art. 
 
Table \ref{tab:comparisons} recalls the performances achieved in the framework of the present paper (columns ``durations'') and the best results given in \cite{Kahindo2018, Kahindo2019} (columns ``velocities''). They were all obtained by cross-validations (LOOCV).

\begin{table}[htp]
\caption{Comparisons between the accuracies of classifications HC/ES-AD achieved from durations measured in mode 4 of tasks 4, 7, 2 and the accuracies achieved with Bayesian approaches from velocities of the pen to perform tasks 1 and 3.}
\begin{center}
\begin{tabular}{|c||c|c|c||c|c|}
\hline
 ~      & durations & durations & durations & velocities & velocities \\
 ~      & SVMs      & GMs        & Algorithm \ref{algo:model} &  Bayes      & Bayes \\
 \hline
task lists: & 2, 4, 7      & 2, 4, 7      & 2, 4, 7     &  1       & 3\\
 \hline
 \hline
error counts & 3 & 7 & 6 & 12 & 14 \\
\hline
accuracies  & 94\% & 87\% & 89\% & 77\% & 74\% \\
\hline
specificities & 93\% & 78\% & 89\%  & 77\% & 72\% \\
\hline
sensitivities & 96\% & 96\% & 88\% & 77\% & 76\% \\
\hline

\end{tabular}
\end{center}
\label{tab:comparisons}
\end{table}%

The results given in Table \ref{tab:comparisons} show that the numbers of errors achieved from the durations of tasks 2, 4, 7 are two,  or even four times lower than those achieved from the velocity measurements carried out for tasks 1 or 3.


\section{Conclusion \label{sec:conclusion}}

This paper suggests a simple approach to recognize ES-AD patients with a high accuracy. It is based on measuring their response  times to perform a specific set of handwriting and drawing tasks.

First, a classification of response times vectors with Support Vector Machines  is used. A high accuracy of 94\% with nested cross-validations was reached, which is by far the best result to date compared to previous works using the same dataset. A grid search associated with the SVMs revealed that the best accuracy is obtained from the durations of three tasks: ``writing a free text'', doing a ``Fitt's test'' and ``holding the tip of a pen in a given position''.

A classification with Gaussian distribution mixtures is then performed to better understand why SVMs perform so efficiently from the three tasks listed above. 
These Gaussian mixtures showed that the distributions of HC and ES-AD samples are well separated by a set of 3 hyperplanes in the feature space.

Thus, a deterministic ad hoc algorithm is derived from the Gaussian model to predict a label HC or ES-AD from the three tasks mentioned above. It reaches 89\% accuracy assessed from cross-validations. In addition, it has the advantage to require the optimization of only 3 parameters during its training stage. The small number of parameters relative to the size of the dataset should significantly reduce the risk of overfitting compared to SVMs.

Considering the best results of the state of the art, the methods presented in this paper divide the number of classification errors by two or four, depending on the learning model chosen.
The Gaussian model suggests that there would be neurological dependencies to be identified between the task durations that characterize the HC or ES-AD classes of samples.

This paper was written from a machine learning perspective. The good results in terms of accuracy are interesting in themselves. However, they raise questions that pertain to the medical field. They should be able to be interpreted from a neurological point of view in order to explain in depth the bases of the observed phenomena.



\section*{Acknowledgments}

This work was funded by {\em Institut Mines-Télécom}, France. The author wishes to thank Prof. Mounim El Yacoubi and Dr. Christian Kahindo for fruitful discussions on different approaches to recognize ES-AD cases. The author is very grateful to Dr. Christian Kahindo for providing the data necessary for this work.


\section*{Competing interests}

The author declares no conflict of interest.


\bibliographystyle{plain}
\bibliography{Alzheimer}

\end{document}